\documentclass[amsmath,amssymb]{revtex4}
\usepackage{graphicx}
\usepackage{bm}

\newcommand{\be}{\begin{eqnarray}}
\newcommand{\ee}{\end{eqnarray}}

\newcommand{\Tr}{{\,\rm Tr\,}}

\newcommand{\ba}{\begin{array}}
\newcommand{\ea}{\end{array}}

\begin{document}
\title{
On Geometric Algebra representation of Binary Spatter Codes}
\author{Diederik Aerts $^1$, Marek Czachor $^{1,2}$, and Bart De Moor $^3$}
\affiliation{
$^1$ Centrum Leo Apostel (CLEA) and Foundations of the Exact Sciences (FUND)\\
Brussels Free University, 1050 Brussels, Belgium\\
$^2$ Katedra Fizyki Teoretycznej i Informatyki Kwantowej\\
Politechnika Gda\'nska, 80-952 Gda\'nsk, Poland\\
$^3$ ESAD-SCD, K. U. Leuven, 3001 Leuwen, Belgium}

\begin{abstract}
Kanerva's Binary Spatter Codes are reformulated in terms of geometric algebra. The key ingredient of the construction is the representation of XOR binding in terms of geometric product.  
\end{abstract}
\maketitle

\section{Introduction}

Distributed representation is a way of representing information in a pattern of activation over a set of neurons, in which each concept is represented by activation over multiple neurons, and each neuron participates in the representation of multiple concepts \cite{E1}.  Examples of distributed representations include Recursive Auto-Associative Memory (RAAM) 
\cite{RAAM}, Tensor Product Representations \cite{Smolensky}, Holographic Reduced Representations (HRRs) \cite{Plate95,Plate2003}, and Binary Spatter Codes (BSC) 
\cite{Kanerva96,Kanerva97,Kanerva98}. 

BSC is a powerful and simple method of representing hierarchical structures in connectionist systems and may be regarded as a binary version of HRRs. Yet, BSC has some drawbacks associated with the representation of chunking. This is why different versions of BSC can be found in the literature. In \cite{Kanerva96,Kanerva97} chunking is given by a majority-rule thresholded addition of binary strings, an operation that often discards a lot of important information. In \cite{Kanerva98}  the ordinary addition is employed, and bits are parametrized differently.

The main message we want to convey in this paper is that there exists a very natural representation of BSC at the level of Clifford algebras. Binding of vectors is here performed by means of the Clifford product and chunking is just ordinary addition. Since Cliford algebras possess a geometric interpretation in terms of Geometric Algebra (GA) 
\cite{GA1,GA2,GA3}, the cognitive structures processed in BSC or HHRs obtain a geometric content. This is philosophically consistent with many other approaches where cognition is interpreted in geometric terms 
\cite{Widdows,PG}. Of particular relevance may be the links to neural computation whose GA and HRR versions were formulated by different authors (cf. \cite{Plate2003,Bayro,JN}. 

The present paper can be also seen in a wider context of a ``quantum structures" approach to cognitive problems we have outlined elsewhere \cite{Gabora,AC,Gabora2,Gabora3,Bart}. 
Cartan's representation of GA in terms of tensor products of Pauli matrices introduces formal links to quantum computation (cf. 
\cite{Somaroo,BDM1,BDM2,BDM3}). 
The philosophy we advocate here is also not that far from the approach of Widdows, where both geometric an ``quantum" aspects play an important role 
\cite{Widdows,WiddowsPeters,WiddowsHiggins}.  

It should be stressed that the GA calculus has already proved to be a powerful tool in applied branches of computer science (computer vision \cite{Lasenby96}, robotics \cite{robot}). GA is a comprehensive language that simplified and integrated many branches of classical and quantum physics \cite{Hestenes}. One may hope that it will play a similar role in cognitive science. 

\section{Binary Spatter Codes}

In BSC information is encoded into long unstructured strings of bits that form a {\it holistic record\/}. The record is composed in two steps called binding and chunking. 

Binding of a role {\bf x} with a filler {\bf y} is performed by means of XOR $\oplus$ (componentwise addition of binary strings mod 2); the role-filler object is ${\bf x}\oplus{\bf y}$. Chunking means adding the bound structures in a suitable way. 

In order to illustrate the original BSC and its algebraic modification let us take the example from \cite{Kanerva97}. The encoded record is
\be
{\bf PSmith}
&=&
{\bf name}
\oplus
{\bf Pat}
+
{\bf sex}
\oplus
{\bf male}
+
{\bf age}
\oplus
{\bf 66}.
\ee
Decoding of the ``name" looks as follows
\be
{\bf Pat'}
&=&
{\bf name}
\oplus
{\bf PSmith}\nonumber\\
&=&
{\bf name}
\oplus
\big[
{\bf name}
\oplus
{\bf Pat}
+
{\bf sex}
\oplus
{\bf male}
+
{\bf age}
\oplus
{\bf 66}
\big]
\nonumber\\
&=&
{\bf Pat}
+
{\bf name}
\oplus
{\bf sex}
\oplus
{\bf male}
+
{\bf name}
\oplus
{\bf age}
\oplus
{\bf 66}
\nonumber\\
&=&
{\bf Pat}
+
{\rm noise}
\to {\bf Pat}.
\ee
We have used here the involutive nature of XOR and the fact that the ``noise" can be eliminated by clean-up memory. The latter means that we compare ${\bf Pat'}$ with records stored in some memory and check, by means of the Hamming distance, which of the stored elements is closest to ${\bf Pat'}$. A similar trick could be done be means of circular convolution in HRRs, but then we would have used an approximate inverse {\bf name}$^*$, and an appropriate measure of distance. Again, the last step is comparison of the noisy object with ``pure" objects stored in clean-up memory.

\section{Geometric-algebra representation of Binary Spatter Codes}

Euclidean-space GA is constructed as follows. One takes an $n$-dimensional linear space with orthonormal basis $\{e_1,\dots,e_n\}$. Directed subspaces are then associated with the set
\be
\{1,e_1,\dots,e_n,e_{12},e_{13}\dots,e_{n-1,n},\dots,e_{12\dots n}\}.
\ee
Here 1 corresponds to scalars, i.e. a 0-dimensional space. Then we have vectors (oriented segments), bivectors (oriented parallelograms), and so on. There exists a natural parametrization: $1=e_{0\dots 0}$, 
$e_1=e_{10\dots 0}$, $e_2=e_{010\dots 0}$, $\dots$, 
$e_{125}=e_{110010\dots 0}$, $\dots$, $e_{12\dots n-1,n}=e_{11\dots 1}$, which shows that there is a one-to-one relation between an $n$-bit number and an element of GA. An element with $k$ 1s and $n-k$ 0s is called a $k$-blade. 

A {\it geometric product\/} of $k$ 1-blades is a $k$-blade. For example, 
$e_{1248}=e_{1}e_{2}e_{4}e_{8}$. Moreover, $e_ne_m=-e_me_n$, if $m\neq n$, and $e_ne_n=1$, for any $n$. GA is a Clifford algebra \cite{BT} enriched by certain geometric interpretations and operations.

Particularly interesting is the form of the geometric product that occurs in the binary parametrization. Let us work out a few examples:
\be
e_1e_1
&=&
e_{10\dots 0}e_{10\dots 0}
=
1=e_{0\dots 0}=e_{(10\dots 0)\oplus (10\dots 0)}\\
e_1e_{12}
&=&
e_{10\dots 0}e_{110\dots 0}
=
e_1e_1e_2=e_2=e_{010\dots 0}=e_{(10\dots 0)\oplus (110\dots 0)}\\
e_{12}e_1
&=&
e_{110\dots 0}e_{10\dots 0}
=
e_1e_2e_1=-e_2e_1e_1=-e_2=-e_{010\dots 0}=-e_{(110\dots 0)\oplus(10\dots 0)}\\
e_{1257}e_{26}
&=&
e_{11001010\dots 0}e_{0100010\dots 0}
=
e_1e_2e_5e_7e_2e_6
=
(-1)^2e_1e_2e_2e_5e_7e_6
=
(-1)^2(-1)^1e_1e_2e_2e_5e_6e_7\nonumber\\
&=&
(-1)^3e_1e_5e_6e_7
=
(-1)^3
e_{10001110\dots 0}
=
(-1)^D
e_{(11001010\dots 0)\oplus(0100010\dots 0)}.
\ee
The number $D$ is the number of times a 1 from the right string had to ``jump" over a 1 from the left one during the process of shifting the right string to the left.
Symbolically the operation can be represented as
\be
\left[
\begin{array}{rl}
\longleftarrow & 01000100\dots 0\\
11001010\dots 0 & 
\end{array}
\right]
\mapsto
(-1)^D
\left[
\begin{array}{l}
01000100\dots 0\\
11001010\dots 0 
\end{array}
\right]
\mapsto
(-1)^D
\left[
\begin{array}{c}
01000100\dots 0\\
\oplus\\
11001010\dots 0 
\end{array}
\right]
=
(-1)^D
\left[
\begin{array}{l}
10001110\dots 0
\end{array}
\right]\nonumber
\ee
The above observations, generalized to arbitrary strings of bits, yield
\be
e_{A_1\dots A_n}e_{B_1\dots B_n}
&=&
(-1)^{\sum_{k<l}B_kA_l}e_{(A_1\dots A_n)\oplus(B_1\dots B_n)}.
\label{GAr}
\ee
Indeed, for two arbitrary strings of bits we have
\be
\left[
\begin{array}{rl}
\longleftarrow & B_1B_2\dots B_n\\
A_1A_2\dots A_n & 
\end{array}
\right]
\mapsto
(-1)^D
\left[
\begin{array}{l}
B_1B_2\dots B_n\\
A_1A_2\dots A_n 
\end{array}
\right]
\ee
where 
\be
D=B_1(A_2+\dots+A_n)+B_2(A_3+\dots+A_n)+\dots+B_{n-1}A_n=\sum_{k<l}B_kA_l.
\ee
We conclude that the map 
\be
(A_1\dots A_n)\times(B_1\dots B_n)
\mapsto
(A_1\dots A_n)\oplus(B_1\dots B_n)
\ee
has GA projective (i.e. up to a sign) representation by means of (\ref{GAr}). Accordingly, the geometric product is a representation of Kanerva's binding at the level of GA. 

Chunking can be represented in GA similarly to what is done in HRRs, that is, by ordinary addition. To see this consider
\be
X &=& e_{A_1\dots A_n}e_{B_1\dots B_n}+e_{C_1\dots C_n}e_{D_1\dots D_n},\\
Y &=&
(-1)^{\sum_{k<l}A_kA_l}e_{A_1\dots A_n} X = e_{B_1\dots B_n}
+
(-1)^{\sum_{k<l}A_kA_l}e_{A_1\dots A_n}e_{C_1\dots C_n}e_{D_1\dots D_n}\\
&=&
e_{B_1\dots B_n}
\pm
e_{(A_1\dots A_n)\oplus(C_1\dots C_n)\oplus(D_1\dots D_n)}\\
&=& e_{B_1\dots B_n}
+{\rm noise}.
\ee
Until now the procedure is similar to what is done in BSC and HRRs. 

An analogue of clean-up memory can be constructed in various ways. One possibility is to make sure that fillers, $e_{B_1\dots B_n}$ etc. are orthogonal to the noise term. For example, let us take the fillers of the form $e_{B_1\dots B_{k}0\dots 0}$, where the first $k\ll n$ bits are selected at random, but the remaining $n-k$ bits are all 0. Let the roles be taken, as in Kanerva's BSC, with {\it all\/} the bits generated at random. The term $e_{(A_1\dots A_n)\oplus(C_1\dots C_n)\oplus(D_1\dots D_n)}$ will with high probability contain at least one $B_j=1$, $k<j\leq n$, and thus will be orthogonal to the fillers. The clean-up memory will consist of vectors with $B_j=0$, $k<j\leq n$, i.e of the filler form. 

Tle final step is performed again in analogy to HRRs. We compute a scalar product between $Y$ and the elements of clean-up memory. Depending on our needs we can play with different scalar products, or with the so-called contractions \cite{Dorst}. The richness of GA opens here  several possibilities. 

\section{Cartan representation}

In this section we give an explicit matrix representation of GA. We begin with Pauli's matrices 
\be
\sigma_1
=
\left(
\begin{array}{cc}
0 & 1\\
1 & 0
\end{array}
\right),
\quad
\sigma_2
=
\left(
\begin{array}{cc}
0 & -i\\
i & 0
\end{array}
\right)
,
\quad
\sigma_3
=
\left(
\begin{array}{cc}
1 & 0\\
0 & -1
\end{array}
\right).
\ee
GA of a plane is represented as follows:  $1=2\times 2$ unit matrix, $e_1=\sigma_1$, 
$e_2=\sigma_2$, $e_{12}=\sigma_1\sigma_2=i\sigma_3$. Alternatively, we can write
$e_{00}=1$, $e_{10}=\sigma_1$, $e_{01}=\sigma_2$, 
$e_{11}=i\sigma_3$, and
\be
\alpha_{00} e_{00}+\alpha_{10} e_{10}+\alpha_{01} e_{01}+\alpha_{11} e_{11}
=
\left(
\begin{array}{cc}
\alpha_{00} +i\alpha_{11} & \alpha_{10} -i\alpha_{01}\\
\alpha_{10} +i\alpha_{01} & \alpha_{00}-i\alpha_{11}
\end{array}
\right).
\ee
This is equivalent to encoding $2^2=4$ real numbers into two complex numbers. 

In 3-dimensional space we have
$1=2\times 2$ unit matrix, $e_1=\sigma_1$, 
$e_2=\sigma_2$, $e_{3}=\sigma_3$, $e_{12}=\sigma_1\sigma_2=i\sigma_3$, 
$e_{13}=\sigma_1\sigma_3=-i\sigma_2$, 
$e_{23}=\sigma_2\sigma_3=i\sigma_1$, 
$e_{123}=\sigma_1\sigma_2\sigma_3=i$.

Now the representation of 
\be
\sum_{ABC=0,1}\alpha_{ABC}e_{ABC}
=
\left(
\begin{array}{cc}
\alpha_{000} +i\alpha_{111} + \alpha_{001}+i \alpha_{110},
& 
\alpha_{100}+i\alpha_{011} -i\alpha_{010}-\alpha_{101}\\
\alpha_{100}+i\alpha_{011} +i\alpha_{010}+\alpha_{101},
& 
\alpha_{000}+i\alpha_{111}
-\alpha_{001}-i \alpha_{110}
\end{array}
\right)
\ee
is equivalent to encoding $2^3=8$ real numbers into 4 complex numbers.

An arbitrary $n$-bit record can be encoded into the matrix algebra known as Cartan's representation of Clifford algebras \cite{BT}:
\be
e_{2k}
&=&
\underbrace{\sigma_1\otimes\dots\otimes \sigma_1}_{n-k}
\otimes\,\sigma_2\otimes
\underbrace{1\otimes\dots\otimes 1}_{k-1},\\
e_{2k-1}
&=&
\underbrace{\sigma_1\otimes\dots\otimes \sigma_1}_{n-k}
\otimes\,\sigma_3\otimes
\underbrace{1\otimes\dots\otimes 1}_{k-1}.
\ee
In practical calculations it is convenient to work with the tensor product implemented by means of the ``drag-and-drop" rule: For arbitrary matrices $A$ and $B$ (not necessarily square, and possibly of different dimensions) 
\be
A\otimes B
&=&
\left(
\begin{array}{ccc}
a_{11} &\dots & a_{1j}\\
\vdots & \ddots & \vdots\\
a_{i1} &\dots & a_{ij}
\end{array}
\right)
\otimes
\left(
\begin{array}{ccc}
b_{11} &\dots & b_{1s}\\
\vdots & \ddots &\vdots\\
b_{r1} &\dots & b_{rs}
\end{array}
\right)
=
\left(
\begin{array}{ccc}
a_{11}B &\dots & a_{1j}B\\
\vdots &\ddots &\vdots\\
a_{i1}B &\dots & a_{ij}B
\end{array}
\right)
=
\left(
\begin{array}{ccc}
a_{11}b_{11} &\dots & a_{1j}b_{1s}\\
\vdots &\ddots &\vdots\\
a_{i1}b_{r1} &\dots & a_{ij}b_{rs}
\end{array}
\right)
.
\ee
This representation of $\otimes$ can be used to show that all $e_k$ given by Cartan's representation are matrices of zero trace. 

\section{Pat Smith revisited}

So let us return to the example from Sec.~2. For simplicity take $n=4$ so that we can choose the representation
\be
\left.
\begin{array}{rcl}
e_{\rm Pat} &=& e_{1100},\\
e_{\rm male} &=& e_{1000},\\
e_{\rm 66} &=& e_{0100},
\end{array}
\right\}{\rm fillers}
\ee
\be
\left.
\begin{array}{rcl}
e_{\rm name} &=& e_{1010},\\
e_{\rm sex} &=& e_{0111},\\
e_{\rm age} &=& e_{1011}.
\end{array}
\right\}{\rm roles}
\ee
The fillers have only the first two bits selected at random, the last two are 00. 
The roles are numbered by randomly selected strings of bits. 

The explicit matrix representations are: 
\be
e_{\rm Pat} &=& e_{1100}=e_1e_2
=
(\sigma_1\otimes\sigma_1\otimes \sigma_1\otimes\,\sigma_3)
(\sigma_1\otimes\sigma_1\otimes \sigma_1\otimes\,\sigma_2)
=
1\otimes 1\otimes 1\otimes\,\sigma_3\sigma_2
=
1\otimes 1\otimes 1\otimes (-i\sigma_1)
\\
e_{\rm male} &=& e_{1000}=e_1
=
\sigma_1\otimes\sigma_1\otimes \sigma_1
\otimes\,\sigma_3\\
e_{\rm 66} &=& e_{0100}
=
e_2
=
\sigma_1\otimes\sigma_1\otimes \sigma_1\otimes\,\sigma_2\\
e_{\rm name} &=& e_{1010}
=
e_1e_3
=
(\sigma_1\otimes\sigma_1\otimes \sigma_1\otimes\,\sigma_3)
(\sigma_1\otimes\sigma_1\otimes\,\sigma_3\otimes 1)\nonumber\\
&=&
1\otimes 1\otimes \sigma_1\sigma_3\otimes\,\sigma_3
=
1\otimes 1\otimes (-i\sigma_2)\otimes\,\sigma_3
\nonumber\\
e_{\rm sex} &=& e_{0111}
=
e_2e_3e_4
=
(\sigma_1\otimes\sigma_1\otimes \sigma_1\otimes\,\sigma_2)
(\sigma_1\otimes\sigma_1\otimes\,\sigma_3\otimes 1)
(\sigma_1\otimes\sigma_1\otimes\,\sigma_2\otimes 1)\nonumber\\
&=&
(\sigma_1\otimes\sigma_1\otimes \sigma_1\otimes\,\sigma_2)
(1\otimes 1\otimes\,\sigma_3\sigma_2\otimes 1)
=
\sigma_1\otimes\sigma_1\otimes \sigma_1\sigma_3\sigma_2\otimes\,\sigma_2
=
\sigma_1\otimes\sigma_1\otimes (-i 1)\otimes\,\sigma_2,\\
e_{\rm age} &=& e_{1011}=e_1e_3e_4
=
(\sigma_1\otimes\sigma_1\otimes \sigma_1\otimes\,\sigma_3)
(\sigma_1\otimes\sigma_1\otimes\,\sigma_3\otimes 1)
(\sigma_1\otimes\sigma_1\otimes\,\sigma_2\otimes 1)\nonumber\\
&=&
\sigma_1\otimes\sigma_1\otimes (-i 1)\otimes\,\sigma_3
\ee
The whole record
\be
{\bf PSmith}
&=&
\alpha
e_{\rm name}e_{\rm Pat}
+
\beta
e_{\rm sex}e_{\rm male}
+
\gamma
e_{\rm age}
e_{\rm 66}\nonumber\\
&=&
\alpha
e_{1010}e_{1100}
+
\beta
e_{0111}e_{1000}
+
\gamma
e_{1011}e_{0100}\nonumber\\
&=&
\alpha
(-1)^{2}
e_{(1010)\oplus(1100)}
+
\beta
(-1)^{3}
e_{(0111)\oplus(1000)}
+
\gamma
(-1)^{2}
e_{(1011)\oplus(0100)}\nonumber\\
&=&
\alpha
e_{0110}
-
\beta
e_{1111}
+
\gamma
e_{1111}\nonumber
\ee
The fact that the last two terms are linearly dependent is a consequence of too small dimensionality of our binary strings (four bits, whereas in realistic cases Kanerva suggested $10^4$ bit strings). This is the price we pay for simplicity of the example. Decoding the name involves two steps. First
\be
e_{\rm name}{\bf PSmith}
&=&
e_{1010}{\bf PSmith}
=
e_{1010}
\big[
\alpha
e_{0110}
-
\beta
e_{1111}
+
\gamma
e_{1111}
\big]\nonumber\\
&=&
\alpha
(-1)^{1}
e_{(1010)\oplus(0110)}
-
\beta
(-1)^{2}
e_{(1010)\oplus(1111)}
+
\gamma
(-1)^{2}
e_{(1010)\oplus(1111)}
\nonumber\\
&=&
-\alpha
e_{1100}
-
\beta
e_{0101}
+
\gamma
e_{0101}
\nonumber\\
&=&
-\alpha
e_{\rm Pat}
\underbrace{
-
\beta
e_{0101}
+
\gamma
e_{0101}}_{\rm noise}={\bf Pat'}.
\ee
It remains to employ clean-up memory. But this is easy since the noise is perpendicular to 
$e_{\rm Pat}$. We only have to project on the set spanned by the fillers, and within this set check which element is closest to the cleaned up ${\bf Pat'}$. 

Cartan's represenation allows to define scalar product in GA by means of the trace. We therefore compare scalar products between ${\bf Pat'}$ and elements of clean-up memory. The only nonzero scalar product is
\be
\langle 
e_{\rm Pat}|{\bf Pat'}\rangle
&=&
\Tr\Big( e_{1100}\big[-\alpha
e_{1100}
-
\beta
e_{0101}
+
\gamma
e_{0101}
\big]\Big)\nonumber\\
&=&
\Tr(-\alpha
e_1e_2e_1e_2
-
\beta
e_1e_2e_2e_4
+
\gamma
e_1e_2e_2e_4
)\nonumber\\
&=&
\Tr(\alpha 1
-
\beta
e_1e_4
+
\gamma
e_1e_4
)=16\alpha.
\ee
Indeed
\be
e_1e_4
&=&
e_{1000}e_{0001}=e_{1001}
=
(\sigma_1\otimes\sigma_1\otimes \sigma_1\otimes\,\sigma_3)
(\sigma_1\otimes\sigma_1\otimes\,\sigma_2\otimes 1)
=
1\otimes 1\otimes (i\sigma_3)\otimes\,\sigma_3
\nonumber\\
&=&
{\rm diag\,}(i,-i,-i,i,i,-i,-i,i,i,-i,-i,i,i,-i,-i,i)
\ee
has zero trace.

\section{Conclusions}

BSC represented at the level of GA maintain the essential element of the original construction, i.e. binding by means of XOR. However, instead of the straightforward map 
\be
({\bf x},{\bf y})\mapsto {\bf x}\oplus{\bf y} 
\ee
we rather have the ``exponential map" ${\bf x}\mapsto e_{\rm x}$ satisfying 
$e_{\rm x}e_{\rm y}=\pm e_{{\rm x}\oplus {\rm y}}$. 
Another difference is in mathematical implementation of chunking. Unbinding produces a noise term which, with high probability, is orthogonal to the original filler. In this respect the construction is analogous to error correcting linear codes. As opposed to tensor product representations, and similarly to BSC and HRRs, binding performed by means of geometric product does not increase dimensions.

\end{document}